\documentclass[sn-basic]{sn-jnl}

\jyear{2021}%

\theoremstyle{thmstyleone}%
\theoremstyle{thmstyletwo}%

\theoremstyle{thmstylethree}%

\usepackage{booktabs}
\usepackage{soul}
\usepackage{amsmath}
\usepackage{amssymb}
\usepackage{makecell}
\usepackage{bbding}
\usepackage{pifont}
\usepackage{bigstrut}
\usepackage{color}
\usepackage{algorithm}
\usepackage{algpseudocode}
\usepackage{multirow}

\begin{document}

\title[Learning Robust Visual-Semantic Embedding for Generalizable Person Re-ID]{Learning Robust Visual-Semantic Embedding for Generalizable Person Re-identification}


\author*[]{\fnm{Suncheng} \sur{Xiang}}\email{xiangsuncheng17@sjtu.edu.cn}

\author[]{\fnm{Jingsheng} \sur{Gao}}\email{gaojingsheng@sjtu.edu.cn}

\author[]{\fnm{Mengyuan} \sur{Guan}}\email{gemini.my@sjtu.edu.cn}

\author[]{\fnm{Jiacheng} \sur{Ruan}}\email{jackchenruan@sjtu.edu.cn}

\author[]{\fnm{Chengfeng} \sur{Zhou}}\email{joe1chief1993@gmail.com}

\author[]{\fnm{Ting} \sur{Liu}}\email{louisa\_liu@sjtu.edu.cn}

\author[]{\fnm{Dahong} \sur{Qian}}\email{dahong.qian@sjtu.edu.cn}

\author[]{\fnm{Yuzhuo} \sur{Fu}}\email{yzfu@sjtu.edu.cn}

\affil[]{\orgname{Shanghai Jiao Tong University}, \orgaddress{\city{Shanghai}, \postcode{200240}, \country{China}}}



%


\abstract{Generalizable person re-identification (Re-ID) is a very hot research topic in machine learning and computer vision, which plays a significant role in realistic scenarios due to its various applications in public security and video surveillance. However, previous methods mainly focus on the visual representation learning, while neglect to explore the potential of semantic features during training, which easily leads to poor generalization capability when adapted to the new domain. In this paper, we propose a \textbf{M}ulti-\textbf{M}odal \textbf{E}quivalent \textbf{T}ransformer called \textbf{MMET} for more robust  visual-semantic embedding learning on visual, textual and visual-textual tasks respectively. To further enhance the robust feature learning in the context of transformer, a dynamic masking mechanism called \textbf{M}asked \textbf{M}ultimodal \textbf{M}odeling  strategy (\textbf{MMM}) is introduced to mask both the image patches and the text tokens, which can jointly works on multimodal or unimodal data and significantly boost the performance of generalizable person Re-ID. Extensive experiments on benchmark datasets demonstrate the competitive performance of our method over previous approaches.  We hope this method could advance the research towards visual-semantic representation learning. Our source code is also publicly available at \url{https://github.com/JeremyXSC/MMET}.}




\maketitle

\section{Introduction}
\label{sec1}
Person re-identification aims to match a specific person in a large gallery with different cameras and locations, which has been studied intensively due to its practical importance in the surveillance system.
With the development of deep convolution neural networks, person Re-ID methods have achieved remarkable performance in a supervised manner~\citep{su2017pose,xiang2020multi}, where a model is trained and tested on different splits of the same dataset.
In practice, however, if we consider each dataset as a domain, there are often huge domain gap since different datasets are often collected in very different visual scenes (\textit{e.g.} indoors, shopping malls, traffic airports). Consequently, the trained models that are directly applied to new domain without model updating are known to suffer from considerable performance degradation~\citep{wang2018transferable, zhong2018generalizing}, thus suggesting model overfitting and poor domain generalization.

In essence, a domain generalizable Re-ID model has great value for real-world large-scale deployment. Specifically, when a customer purchases a Re-ID system for a specific camera network, the system is expected to work out-of-the-box, without the need to go through the tedious process of data collection, annotation and model updating.
Surprisingly, there is very little prior study of this topic. Existing Re-ID works mainly evaluate their models' cross-dataset generalization, but no specific design is made to make the models more generalizable. Recently, unsupervised domain adaptation (UDA) methods for Re-ID have been studied to adapt a Re-ID model from source to target domain~\citep{xiang2020unsupervised,zhang2021unsupervised}. However, previous UDA models update using unlabeled target domain data, which is infeasible for real-world scenarios.

Beyond domain adaptation, the problem of domain generalization (DG) has been investigated in deep learning, with some recent few-shot meta-learning approaches also adapted for domain generalization. However, existing domain generalization methods~\citep{khosla2012undoing,li2018learning,muandet2013domain} assume that the source and target domain have the same label space; whilst existing meta-learning models assume a fixed number of classes for target domains and are trained specifically for that number using source data. They thus have limited efficacy for generalizable Re-ID, where target domains have a different and variable number of identities.
\begin{figure*}[!t]
\centering{\includegraphics[width=0.7\linewidth]{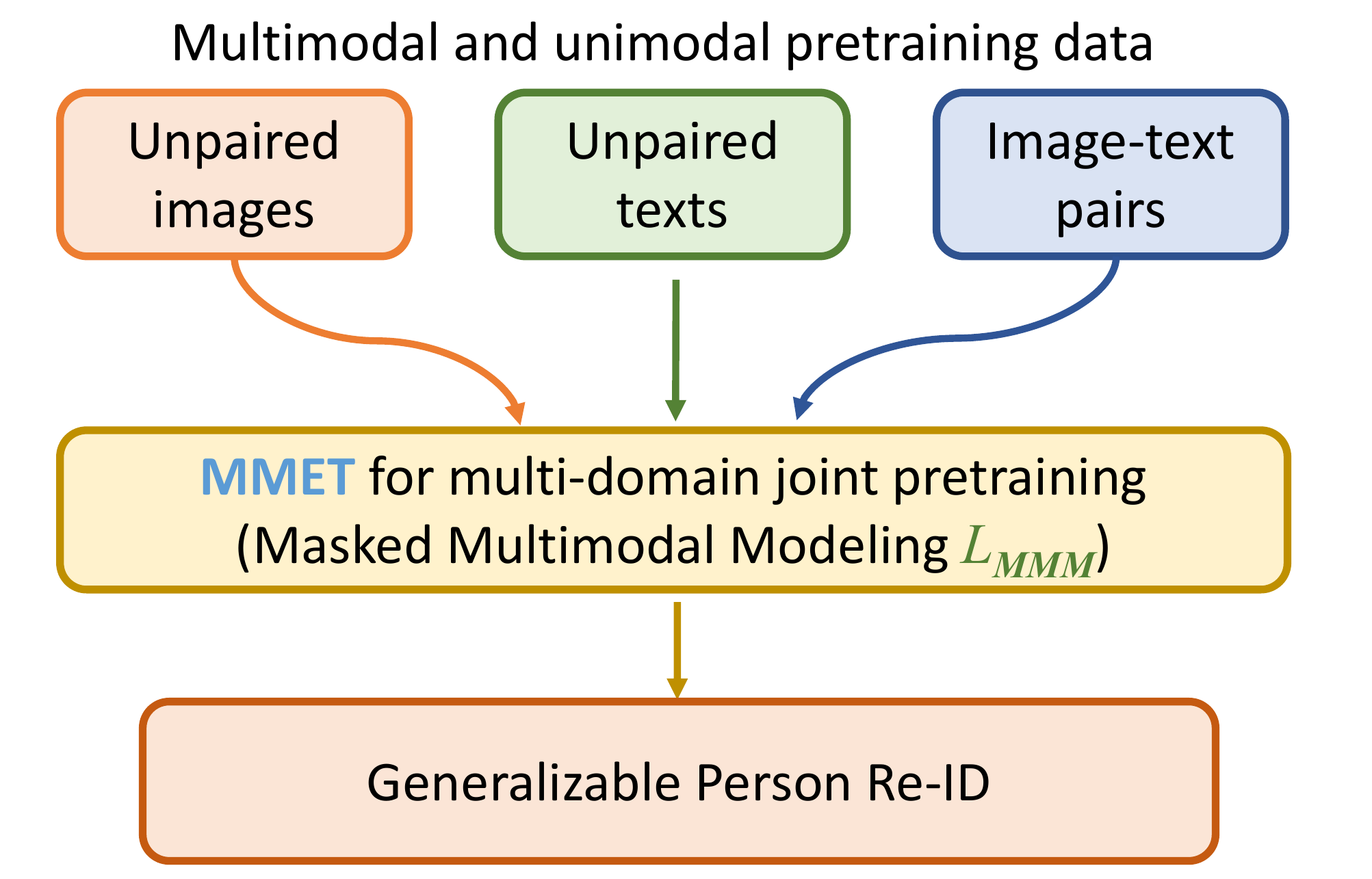}}
\caption{We present MMET, a language and vision alignment model that learns strong representations from image-text pairs and unpaired images / texts under a common transformer model architecture.}
\label{fig1}
\end{figure*}
Besides, these approaches are hardly applicable in practice due to expensive labeling costs and also suffer from severe performance degradation on an
unseen target domain, which is mainly caused by the domain gap between different backgrounds, camera angles, and camera styles, is now becoming the main challenges for Re-ID community.
In common sense, combining information from different modalities into universal architecture holds promise not only because it is similar to how humans make sense of the world, but also because it may lead to better sample efficiency and much richer representations.

In this paper, we aim to learn a domain generalizable Re-ID model on the basis of visual-semantic feature, even in a scenario with incomplete modality (\textit{e.g.} unimodal vision-only or language-only data), and propose a Multi-Modal Equivalent Transformer (MMET) for more robust visual-semantic embedding learning for Re-ID event. On the basis of it, a dynamic masking mechanism called Masked Multimodal Modeling (MMM) is introduced to mask both the image patches and the text tokens, which can jointly works on unimodal visual or textual modality and significantly boost the performance of generalizable person re-identification. As illustrated in Fig.~\ref{fig1}, our model is trained on a set of source domain based on the unpaired images / texts or image-text pairs, and should generalize to any new unseen datasets for effective Re-ID without any model updating.
To the best of our knowledge, this is the first work aiming at investigating the domain generalization with multiple modalities on Re-ID events. Compared with traditional multimodal feature fusion strategies, our MMET is different from them in terms of two perspective: 1) The input of traditional method is unimodal vision-only or language-only data, while our MMET model works on a wide range of tasks in the vision, language, and vision-and-language domains, which allows our model to be more flexible and adaptable in real-world scenario;
2) Our MMET model is designed to be able to take advantage of unpaired image data or text data along with paired image-text pairs. While previous methods fail to adapt in the multimodal scenarios, especially for those cross-modal and multimodal vision-and-language tasks.

As a consequence, the major contributions of our work can be summarized into three-fold:
\begin{itemize}
 \item[$\bullet$] We propose a Multi-Modal Equivalent Transformer framework called MMET for more robust visual-semantic embedding learning.

 \item[$\bullet$] Based on it, a dynamic masking mechanism called Masked Multimodal Modeling is introduced to further boost the performance of generalizable person re-identification.

 \item[$\bullet$] Comprehensive experiments show that our MMET method matches or exceeds the performance of existing methods with a clear margin, which reveals the applicability of visual-semantic based pretraining with new insights.
\end{itemize}

In the rest of the paper, we first review some related works of person re-identification methods and previous semantic-based method in Section \ref{sec2}. Then in Section \ref{sec3}, we give more details about the learning procedure of the proposed MMET method. Extensive evaluations compared with state-of-the-art methods and comprehensive analyses of the proposed approach are elaborated in Section \ref{sec4}. Conclusion and Future Works are given in Section \ref{sec5}.

\section{Related Works}
\label{sec2}
In this section, we have a brief review on the related works of traditional person Re-ID methods and Transformer-based  approaches. The mainstream idea of the existing methods is to learn a robust model for feature representation.

\subsection{CNN-based Person Re-ID Methods}
Actually, there are mainly two kinds of feature learning paradigms for person Re-ID tasks: (1) Hand-crafted based method and (2) Deep learning based approach, which are introduced as follow:

Traditional research works~\citep{farenzena2010person,zhao2014learning,ding2015deep} related to hand-crafted systems for person Re-ID aim to design or learn discriminative representation or pedestrian features. For example, \citep{farenzena2010person} proposed an appearance-based method for these situations where the number of candidates varies continuously.
\cite{ding2015deep} presented a scalable deep feature learning model for person re-identification via distance comparison. Besides directly using mid-level color and texture features, some methods~\citep{zhao2014learning} also explore different discriminative abilities of local patches for better discriminative power and generalization ability.
Unfortunately, these hand-crafted feature based approaches always fail to produce competitive results on large-scale datasets. The main reason is that these early works are mostly based on heuristic design, and thus they could not learn optimal discriminative features on current large-scale dataset.

Recently, benefited from the advances of deep neural networks and availability of large-scale datasets, person Re-ID performance in supervised learning has been significantly boosted to a new level~\citep{xiang2020multi,xiang2021less}, e.g. \cite{xiang2020multi} propose a feature fusion strategy based on traditional convolutional neural network with attention mechanism, which learns robust feature extraction and reliable metric learning in an end-to-end manner. \cite{gu2022clothes} propose a clothes-based adversarial loss to mine clothes-irrelevant features from the original RGB images by penalizing the predictive power of Re-ID model. \cite{wang2022nformer} design a neighbor transformer network to explicitly model interactions across all input images for discriminative representations.
Besides, some recent works~\citep{wei2018person,deng2018image} attempt to address unsupervised domain adaptation base on Generative Adversarial Network (GAN) model.
Unfortunately, these approaches always require abundant computing resources to achieve satisfactory performance, and leveraging  GAN  network is unable to guarantee the quality of generated images.

\subsection{Transformer-based Methods}

Transformer model is proposed in~\citep{vaswani2017attention} to handle sequential data in the field of natural language processing, which has been applied into computer vision to explore long-range dependencies with multi-head self-attention strategy on person Re-ID task. For example, \cite{he2021transreid} make the first attempt to apply Transformer architecture on this event, and propose a pure transformer-based object Re-ID framework with critical improvement. \cite{liao2021transmatcher} also adopt vision transformer for efficient image matching and metric learning tasks. As for the robust feature extraction,
\cite{chen2022rest} design a hybrid backbone Res-Transformer based on ResNet-50 and Transformer block for effective identify information. \cite{lai2021transformer} propose a adaptive part division model to better extract local features for person Re-ID. In addition, \cite{xiang2022deep} propose a deep multimodal fusion network to elaborate rich semantic knowledge for assisting in representation learning during the pretraining. Although these methods can improve the performance of generalizable person Re-ID in some degree, the potential of deep multimodal feature between visual and semantic feature is always being underestimated. Consequently, current methods are still far from satisfactory in generalization for practical person re-identification.

To solve these problems, in this work, we take a big step forward and propose a Multi-Modal Equivalent Transformer (MMET) for more robust visual-semantic embedding learning on visual, textual and visual-textual tasks. On the basis of it, a dynamic masking mechanism called Masked Multimodal Modeling (MMM) is introduced to mask both the image patches and the text tokens, which can jointly work on multimodal or unimodal data and significantly boost the performance of generalizable person Re-ID. To the best of our knowledge, this is the first attempt to adopt a foundational language and vision alignment model that explicitly targets vision, language, and their multimodal combination all at once. We hope that our method will serve as a strong baseline for visual-semantic embedding, and shed light into potential tasks for the community to move forward.

\section{Methodology}
\label{sec3}

\subsection{Problem formulation}
We begin with a formal description of the domain generalizable person re-identification (DG Re-ID) problem. We assume that we are given $K$ source domains $\mathcal{D}=\left\{\mathcal{D}_k\right\}_{k=1}^K$. Each source domain contains its own image-label pairs $\mathcal{D}_k=\left\{\left(\boldsymbol{x}_i^k, y_i^k\right)\right\}_{i=1}^{N_k}$, where  $\mathcal{N}_k$ is the number of images in the source domain $ \mathcal{D}_k$. Each sample $\boldsymbol{x}_i^k \in \mathcal{X}_k$ is associated with an identity label $y_i^k \in \mathcal{Y}_k=\left\{1,2, \ldots, M_k\right\}$, where $\mathcal{M}_k$ is the number of identities in the source domain $\mathcal{D}_k$. In the training phase, we train a DG model using the aggregated image-label pairs of all source domains. In the testing phase, we perform a retrieval task on unseen target domains without additional model updating.

\subsection{Our Proposed MMET framework}
\begin{figure*}[!t]
\centering{\includegraphics[width=1.00\linewidth]{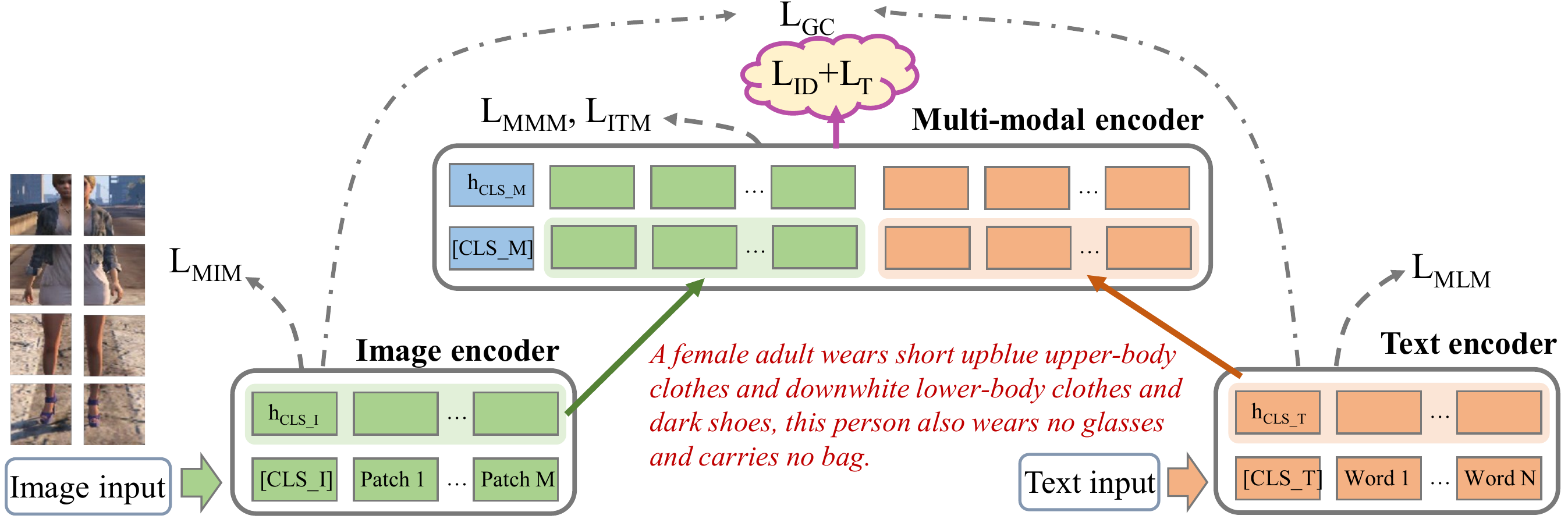}}
\caption{The framework of our proposed MMET method, which contains: 1) \textbf{Image encoder} to capture unimodal image representations; 2) \textbf{Text encoder} to process unimodal text information; 3) \textbf{Multimodal encoder} that takes as input the encoded unimodal image and text and integrates their representations for multimodal reasoning. During pretraining, masked image modeling $\mathcal{L_{MIM}}$ and masked language modeling $\mathcal{L_{MLM}}$ losses are applied onto the image and text encoders over a single image or a text piece, respectively, while masked multimodal modeling $\mathcal{L_{MMM}}$ and image-text matching $\mathcal{L_{ITM}}$ are used over paired image-text data.}
\label{fig2}
\end{figure*}

In this section, we propose a Multi-Modal Equivalent Transformer network (MMET) for multimodal feature learning, which can be flexibly adapted to the visual task, textual task and visual-textual task, respectively. In fact, traditional single-stream structure requires early fusion of the two modalities, which can not be directly adapted in the scenario where the attributes of the benchmark dataset is missing. To address this problem, we adopt the two-stream architecture as the backbone for generalizable person Re-ID event, which contains \textbf{Image encoder}, \textbf{Multi-modal encoder} and \textbf{Multi-modal encoder}, the detailed architecture of our approach is illustrated in Fig.~\ref{fig2}.

\textbf{Image Encoder}
The image encoder mainly extracts visual model features from two-dimensional image input, and the structure still adopts the basic construction of visual transformers. Given an input image, we resize it to a fixed size, and divide the image into $M$ mean image patches, which are then linearly projected through
linear projection layer, with bit-set padding and an extra image classification token [CLS\_I] is sent to the Transformer layer. The output of the image encoder is a sequence of hidden state vectors, which can be denoted as set $\{ h_{I} \}$. Each of these corresponds to an image patch, and there is an additional global feature output $h_{CLS\_I}$ corresponding to the image classification token [CLS\_I] .

\textbf{Text Encoder}
In order to ensure the overall balance of the MMET model, the text encoder here is different from the general model in the field of natural language processing. Instead, we basically adopt the same architecture as the visual Transformer, only the parameters are different. Giving the caption from a pedestrian, we first tokenize the text according to the word sentence. Continuously, we embed it into the word vector sequence according to the BERT model~\citep{devlin2018bert}, then we apply a transformer model over the word vectors to encode them into
a list of hidden state vectors $\{h_{T}\}$, including $ h_{CLS\_T} $ for the text classification  [CLS\_T] token.

\textbf{Multi-modal encoder}
Different from the general two-stream structure, in this work, we adopt a separate Transformer to fuse the hidden state sequence of images and text, as shown in Fig.~\ref{fig2}. Specifically, we apply two linear maps to each hidden state vector in the sets  of $\{h_{I}\}$ and $\{h_{T}\}$, which are then concatenated into a full sequence while adding an additional token [CLS\_M].  As shown in Fig.~\ref{fig2}, this concatenated list is fed into the multimodal encoder transformer, allowing cross-attention between the projected unimodal image and
text representations and fusing the two modalities.
It is worth mentioning that the output value of this part is also a set of hidden state sequences ${h_{M}}$, each of these corresponds to a unimodal vector from $h_{I}$ or $h_{T}$, and a vector $ h_{CLS, M} $ corresponding to [CLS\_M].

\textbf{Network updating}
In this work, we pretrain the entire model from scratch with synthetic dataset FineGPR~\citep{xiang2021less}, which contains more than two million image-text pairs. For the generalizable person Re-ID task, we apply the classifier head on top of the multimodal $h_{CLS, M}$ from the multi-modal encoder during training phase. Similarly, for visual recognition and language understanding tasks, we apply a classifier head on top of $h_{CLS, I}$ from image encoder and $h_{CLS, T}$ from text encoder, respectively. During the model optimization phase, we pretrain the MMET model once, and evaluate it separately on each downstream Re-ID task.
For a single-label $N$ classification task, the identification loss (cross-entropy loss) is written as,
\begin{equation}\label{eq3}
\mathcal{L}_{ID}=-\frac{1}{\text { $M_{batch}$ }} \sum_{i=1}^{\text { $M_{batch}$}} \sum_{j=1}^N y_{i j} \log \hat{y_{i j}}
\end{equation}
where $M_{batch}$ is the number of labeled training images in a batch, $\hat{y_{i j}}$ is the predicted probability of the input belonging to ground-truth class $y_{i j}$.
In essence, many previous works~\citep{xiang2022learning,luo2019bag} have been found that performing training with multiple losses has great potential to learn a robust and generalizable Re-ID model,
Inspired from this, we also adopt triplet loss to mine the relationship of training samples during training, which can minimize the distance among positive pairs and maximize the distance between negative pairs. And our triplet loss is defined as:
\begin{equation}
\mathcal{L}_{triplet}=\left(d_{a, p}-d_{a, n}+m\right)_{+}
\label{eq2}
\end{equation}
where $d_{a, p}$, $d_{a, n}$ denote the feature distances of positive pair and negative pars, respectively, $m$ represents the margin of our triplet loss, $(z)_{+}$ denotes \textit{max(z,0)}.

To this end, we propose to jointly learn robust visual-semantic embedding using classification loss~\citep{zheng2017discriminatively} and Triplet loss~\citep{hermans2017defense} in a training batch, which can be expressed as:

\begin{equation}
\mathcal{L}_{total}= \mathcal{L}_{ID} + \mathcal{L}_{Triplet}
\label{eq1}
\end{equation}

\begin{figure*}[!t]
\centering{\includegraphics[width=1.00\linewidth]{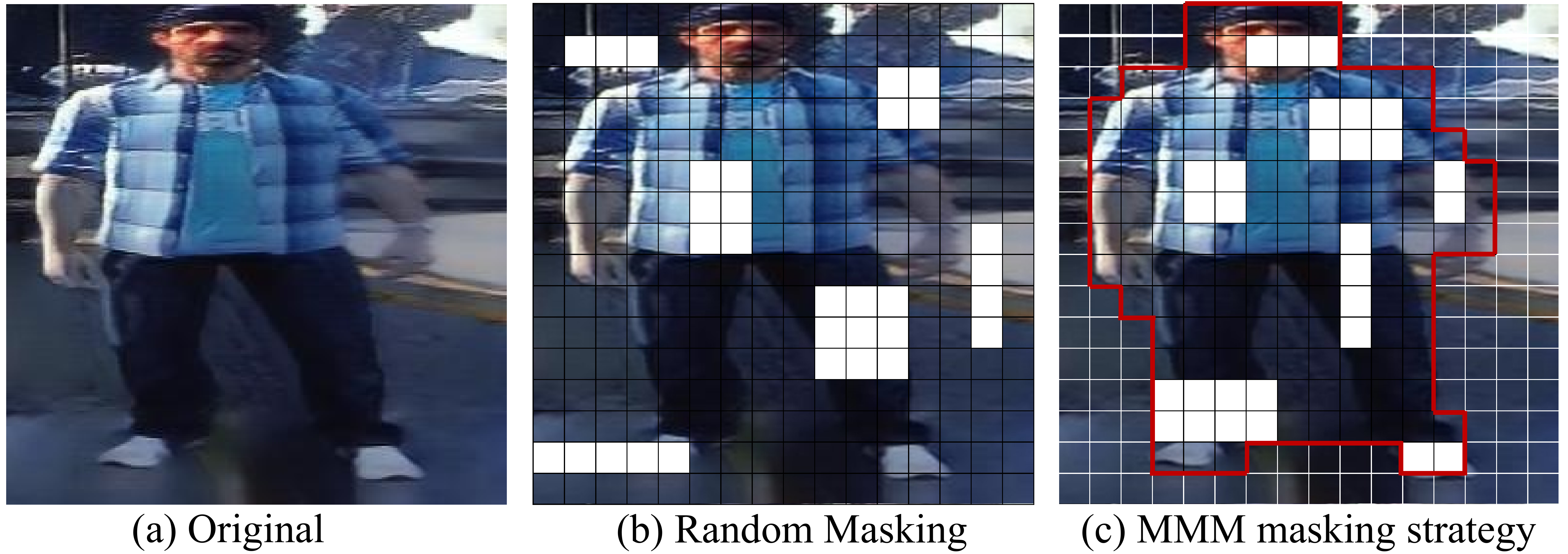}}
\caption{An example of masked image of our masked modality modeling strategy, in which we mainly mask the pedestrian region instead of whole part of input images.}
\label{fig3}
\end{figure*}

\subsection{Masked multimodal modeling strategy}
In fact, most of the previous visual-language modeling approaches focus on the masked language modeling of the texts in multimodal input by reconstructing the mask labeling unit, while neglects the mask feature learning at the pixel level for single image modality. In this work, we introduce a \textbf{M}asked \textbf{M}ultimodal \textbf{M}odeling strategy to perform pretraining based on the MMET model, which can effectively achieve the goal of performing mask loss calculations on image and text modality. More details are depicted in Fig.~\ref{fig3}.

More specifically, we mainly mask the pedestrian region instead of whole part of input images, since the human part contains more discriminative feature for representation learning, and it has no effect on the prediction performance of the model when the masked region mainly focus on the background of input samples. In this work, masked multimodal modeling is a dynamic strategy which can be applied to pretraining tasks on text encoders for monolingual datasets. Following the default settings in the field of natural language processing task~\citep{devlin2018bert}, here we also randomly mask a certain fraction (\textit{e.g.} \textbf{masking ratio: 15\%}) of image blocks and text labeling units respectively.
Then we adopt the classifier to reconstruct from other image blocks and tokenized units on the hidden state output ${h_{M}}$ of the image-text modality. In contrast to denoising auto-encoders~\citep{vincent2008extracting}, we only predict the masked image and words rather than reconstructing the entire text inputs.

During the pretraining, masked image modeling is applied into the image encoder over unpaired image data, while masked language modeling is applied into the text encoder over unpaired text data. Specifically, given a image and text data, we perform masking on image encoder and text encoder respectively.
Finally, we add an image-text matching loss $\mathcal{L_{ITM}}$ following prior vision-and-language pretraining literature~\citep{lu2019vilbert}. On the basis of it, we introduce Image-Text Matching pretraining task to learn the relationship between image and text samples, and then apply a classifier to decide if an input image and text match each other on the basis of ${h_{CLS, M}}$ from the multimodal encoder.
It is worth mentioning that masked image modeling loss $\mathcal{L_{MIM}}$ and masked language modeling loss $\mathcal{L_{MLM}}$ are applied onto the image and text encoders over a single image or a text piece, respectively. While masked multimodal modeling loss $\mathcal{L_{MMM}}$ are used over paired image-text data.

For downstream tasks, classification heads are applied on the outputs from the image, text and multimodal encoders respectively for image retrieval tasks.
According to the output ${h_{M}}$ of the multimodal encoder, a multi-layer perception is applied to perform predictive computation on masked partial image patches and text labeling units.
For the loss function, $\mathcal{L_{MMM}}$ can be regarded as a relatively complete mask training target in the field of visual language multimodality because it combines the mask of images and text at the same time. In fact, our $\mathcal{L_{MMM}}$ loss can be seen as a supplement to the general global contrastive loss $\mathcal{L_{GC}}$.

To sum up, our Multi-Modal Equivalent Transformer framework is expected to be applicable to multimodal or unimodal scenarios for various tasks, which has the great potential to solve the incomplete modality for generalizable  Re-ID event. Besides, masked multimodal modeling strategy can mask the pedestrian region instead of whole part of input images, which can significantly enhance the robustness of visual-semantic embedding learning. The reason for choosing Transformer in this work is not only because it represents a new trend of traditional vision tasks, but also considering that Transformer has its own architectural advantages in the use of more modalities.

\section{Experimental Results}
\label{sec4}
\subsection{Datasets}
\label{sec4.1}
In this paper, we conduct experiments on several large-scale
public datasets, which include Market-1501~\citep{zheng2015scalable}, DukeMTMC-reID~\citep{ristani2016performance,zheng2017unlabeled} and CUHK03~\citep{li2014deepreid} datasets, as well as multimodal Re-ID dataset CUHK-PEDES~\citep{li2017person} and FineGPR~\citep{li2017person}.

\textbf{Market-1501}~\citep{zheng2015scalable} contains 32,668 labeled images of 1,501 identities captured from campus in Tsinghua University.  Each identity is captured by at most 6 cameras. The training set contains 12,936 images from 751 identities and the test set contains 19,732 images from 750 identities.

\textbf{DukeMTMC-reID}~\citep{ristani2016performance,zheng2017unlabeled} is collected from Duke University with 8 cameras, it has 36,411 labeled images belonging to 1,404 identities and contains 16,522 training images from 702 identities, 2,228 query images from another 702 identities and 17,661 gallery images.

\textbf{CUHK03}~\citep{li2014deepreid} contains 14,097 images of 1,467 identities.
Following the CUHK03-NP protocol~\citep{zhong2017re}, it is divided into 7,365 images of 767 identities as the training set, and the remaining 6,732 images of 700 identities as the testing set.

\textbf{CUHK-PEDES}~\citep{li2017person} contains 40,206 images of 13,003 persons from five existing person re-identification datasets, as the subjects for language descriptions. And each image was annotated with two sentence descriptions and a total of 80,412 sentences were collected.

\textbf{FineGPR}~\citep{li2017person} contains 2,028,600 synthesized person images of 1,150 identities. Images in this dataset generally contain different attributes in a large scope, e.g., Viewpoint, Weather, Illumination, Background and ID-level annotations, also including many hard samples with occlusion.

In our experiments, we follow the standard evaluation protocol~\citep{zheng2015scalable} used in Re-ID task , and adopt mean Average Precision (mAP) and Cumulative Matching Characteristics (CMC) at Rank-1 and Rank-5 for performance evaluation on downstream Re-ID task.

\subsection{Implementation Details}
During the pretraining of MMET model,  we take the ViT-B/16~\citep{dosovitskiy2020image} as the backbone for the image encoder, text encoder and multi-modal encoder respectively.
Following the training procedure in~\citep{singh2022flava}, we adopt SGD with momentum of 0.9 and weight decay of 0.1. The learning rate is initialized as 1e-3 with cosine learning rate decay. Additionally,
all the images are resized to the $256 \times 128$, and the length of the segmented image block is $12 \times 12$. The batch size of training samples is set as 64. As for triplet selection, we randomly select 16 persons and sampled 4 images for each identity.  Adam method and warmup learning strategy are also adopted to optimize the model.  All the experiments are performed on PyTorch~\citep{paszke2019pytorch} with one Nvidia GeForce RTX 3090 GPU on a server equipped with a Intel Xeon Gold 6240 CPU.

\begin{table}[!t]
  \centering
  \caption{Ablation studies of masked multimodal modeling on the basis of MMET method. And MMM indicates the masked multimodal modeling strategy proposed in this work.}
  \setlength{\tabcolsep}{3.21mm}{
    \begin{tabular}{lcccc}
    \toprule
    \multirow{2}[4]{*}{Methods} & \multicolumn{2}{c}{Market-1501} & \multicolumn{2}{c}{DukeMTMC-reID} \\
\cmidrule{2-5}          & mAP   & Rank-1 & mAP   & Rank-1 \\
    \midrule
    Baseline (Original) & 15.4  & 38.9  & 9.8   & 19.5 \\
    MMET (Random Masking)  & 26.7  & 51.1  & 19.2  & 35.1 \\
    MMET+MMM (Ours) & 28.8  & 59.2  & 23.5  & 43.6  \\
    \bottomrule
    \end{tabular}}%
  \label{tab2}%
\end{table}%


\subsection{Ablation study}
\textbf{MMET is a strong baseline for visual-semantic embedding.}
In this section, we further validate the effectiveness of our masked multimodal modeling strategy. We give some results of our MMET model on Market-1501 and DukeMTMC-reID dataset respectively. The detailed results are reported in Table~\ref{tab2}. Compared with the baseline model, our proposed MMET can increase the mAP performance on two benchmark datasets from 15.4\%, 9.8\% to 26.7\% (\textbf{+11.3\%}), 19.2\% (\textbf{+9.4\%}), respectively. This indicates that Multi-Modal Equivalent Transformer increases the discriminative ability of the feature. The structure of MMET is as concise as that of Vision Transformer (VIT-B/16), and training MMET requires nothing more than training a canonical classification network. We hope it will serve as a baseline for multimodal image retrieval task.

\textbf{Masked multimodal modeling improves MMET especially in Rank-1 accuracy.}
According to the Table~\ref{tab2}, while MMET already has a high accuracy, MMM brings further improvement to it. On these two datasets, the improvement in rank-1 accuracy is \textbf{+8.1\%} and \textbf{+8.5\%} respectively; the improvement in mAP is \textbf{+2.1\%} and \textbf{+4.3\%} respectively. The improvement in rank-1 is larger than in mAP accuracy. In fact, rank-1 accuracy characterizes the ability to retrieve the easiest match in the camera network, while mAP indicates the ability to find all the matches. So the results indicate that MMM strategy is especially beneficial in finding more closer and easier matches at the first sight, which is more applicable in real-world scenarios.

\begin{table}[!t]
  \centering
  \caption{Validation experiment results on CUHK-PEDES dataset, which contains rich multimodal labels (\textit{e.g.} Image and Text) for multimodal fusion. MIM, MLM and MMM represent the masked image modeling, masked language modeling and masked multimodal modeling strategy respectively.}
   \setlength{\tabcolsep}{1.51mm}{
    \begin{tabular}{lccccc}
    \toprule
    \multicolumn{1}{c}{\multirow{2}[4]{*}{Methods}} & \multicolumn{1}{c}{\multirow{2}[4]{*}{Image data}} & \multirow{2}[4]{*}{Text data} & \multicolumn{3}{c}{FineGPR $\rightarrow$ CUHK-PEDES } \\
\cmidrule{4-6}          &       &       & mAP   & Rank-1 & Rank-5 \\
    \midrule
    MMET \textit{w/} MIM  & $\checkmark$     & $\times$     & 24.6  & 36.7  & 47.6 \\
    MMET \textit{w/}  MLM  & $\times$     & $\checkmark$     & 14.4  & 27.8  & 39.5 \\
    MMET  \textit{w/}  MMM  & $\checkmark$     & $\checkmark$     & 25.6  & 38.8  & 50.4  \\
    \bottomrule
    \end{tabular}}%
  \label{tab3}%
\end{table}%

\textbf{The effectiveness of masked multimodal modeling strategy.} According to the Table~\ref{tab3}, we can obviously observe that the performance of Image \& Text (MMM strategy) is much superior than single modality (\textit{e.g.} Image or Text) (MIM or MLM strategy) , which demonstrates that there exists a mutual benefits between images and text data. It is worth mentioning that CUHK-PEDES dataset contains rich multimodal labels (\textit{e.g.} Image and Text), which can serve as a strong foundation for our visual-semantic embedding learning.
However, from the Table~\ref{tab2} and Table~\ref{tab3}, we can also observe that the performance of our model on CUHK-PEDES dataset is not so competitive than Market-1501 dataset  (mAP \textcolor[rgb]{1.00,0.39,0.09}{\textbf{28.8\%}} vs. \textcolor[rgb]{0.20,0.40,0.80}{\textbf{25.6\%}}), we suspect that this is due to the previously mentioned domain gap in semantic caption between FineGPR and CUHK-PEDES datasets, which hinders the further improvement of our method on CUHK-PEDES dataset.

\begin{table}[!t]
  \centering
  \caption{Comparison with SOTA unsupervised methods on Market-1501, DukeMTMC-reID and CUHK03 respectively. \textbf{Bold} indicates the best and \underline{underline} the second best}
  \setlength{\tabcolsep}{1.2mm}{
    \begin{tabular}{lcccccc}
    \toprule
    \multirow{2}[4]{*}{Methods}  & \multicolumn{2}{c}{Market-1501} & \multicolumn{2}{c}{DukeMTMC-reID} & \multicolumn{2}{c}{CUHK03} \\
\cmidrule{2-7}                & mAP   & Rank-1 & mAP   & Rank-1 & mAP   & Rank-1 \\
    \midrule
    LOMO~\citep{liao2015person}     & 8.0  & 27.2   & 4.8     & 12.3    & -     & -  \\
    BOW~\citep{zheng2015scalable}      & 14.8  & 35.8   & 8.3     & 17.1    & -     & -  \\
    \midrule
    PTGAN~\citep{wei2018person}     & 15.7  & 38.6   & 13.5     & 27.4    & -     & -  \\
    SPGAN~\citep{deng2018image}      & 22.8  & 51.5   &  \underline{22.3}     &  \underline{41.1}    & -     & -  \\
    QAConv~\citep{liao2020interpretable}  &  \underline{27.2}  &  \underline{58.6}  & -     & -     & 6.8   & 7.9 \\
    QAConv-GS~\citep{liao2022graph}   & -     & -     & -     & -     &  \underline{15.7}  &  \underline{16.4} \\
    CDTnet~\citep{guan2022cdtnet}   & 17.8  & 35.6  & 12.3  & 27.9  & 10.9  & 11.3 \\
    \midrule
    \textbf{MMET \textit{w/} MMM (Ours)}  & \textbf{28.8}  & \textbf{59.2}  & \textbf{23.5}  & \textbf{43.6}  & \textbf{20.3}  & \textbf{23.6} \\
    \bottomrule
    \end{tabular}}%
  \label{tab1}%
\end{table}%


\subsection{Comparison to the state-of-the-art models}
In this section, we compare our proposed method with several state of the art Re-ID backbones. which can be divided into hand-crafted feature approaches: LOMO~\citep{liao2015person} and BOW~\citep{zheng2015scalable}; deep learning-based methods: PTGAN~\citep{wei2018person}, SPGAN~\citep{deng2018image}, QAConv~\citep{liao2020interpretable}, QAConv-GS~\citep{liao2022graph} and CDTnet~\citep{guan2022cdtnet}. More detailed comparison results of these methods are demonstrated as follows:

According to the results in Table~\ref{tab1}, our proposed method surpasses all the prior methods and can achieve the competitive performance on Market-1501, DukeMTMC-reID and CUHK03 dataset respectively. For example, we can obtain a remarkable mAP performance of 28.8\% on Market-1501 dataset, leading a significant improvement of \textbf{+1.6\%} when compared with second best method QAConv~\citep{liao2020interpretable}. Similarly, we can also arrive at rank-1 accuracy= 43.6\% and 23.6\% on DukeMTMC-reID and CUHK03 dataset respectively, higher than the second best methods of SPGAN~\citep{deng2018image} and QAConv-GS~\citep{liao2022graph} separately. These comparisons indicate the superiority and competitiveness of the proposed MMET method on standard Re-ID benchmarks.

\subsection{Visualization results}
\begin{figure*}[!t]
\centering{\includegraphics[width=1.00\linewidth]{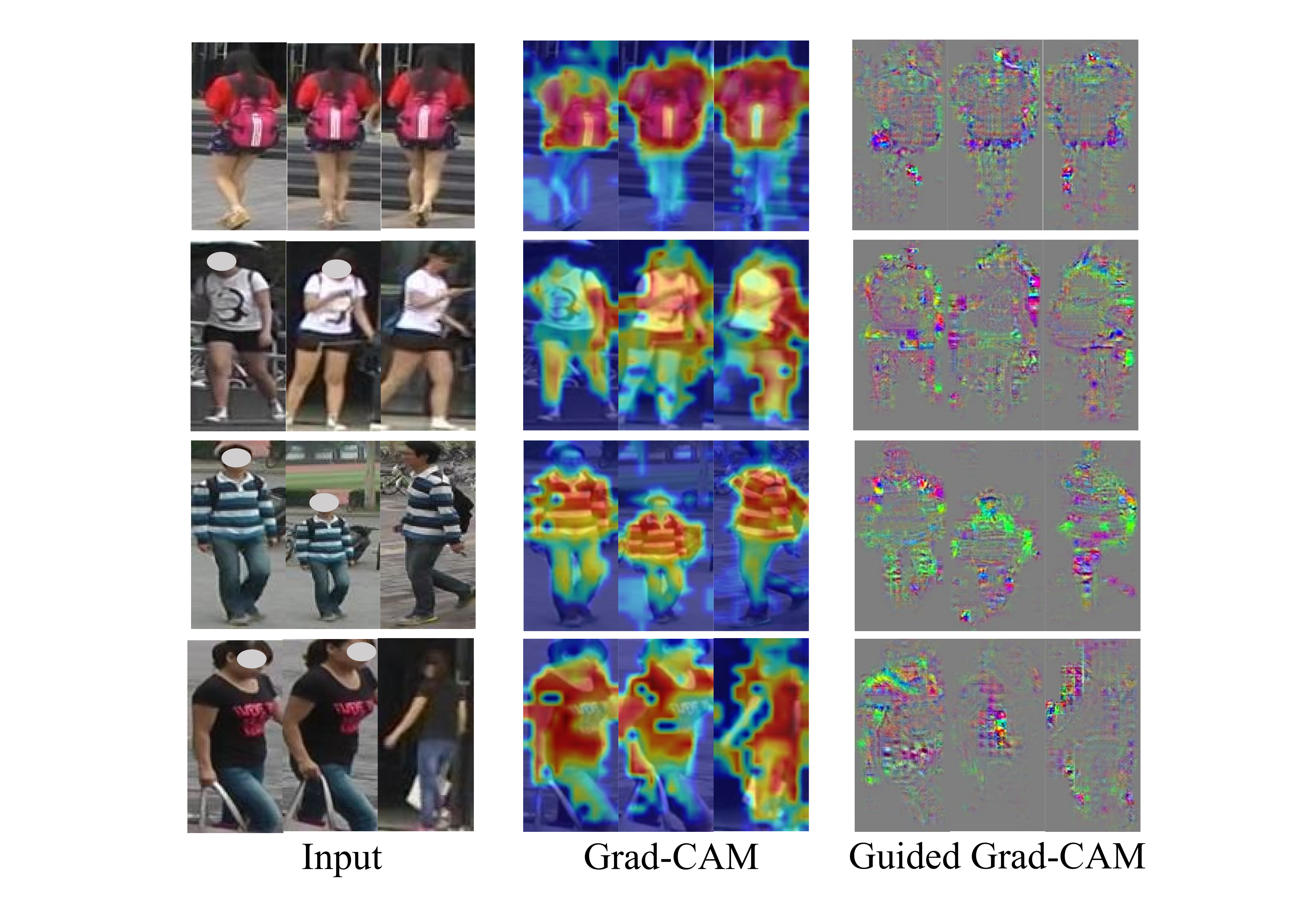}}
\caption{Visualization results of the Grad-CAM attention map of the proposed MMET model: (a) Original image; (b) Grad-CAM; (c) Guided Grad-CAM.}
\label{fig4}
\end{figure*}

To further demonstrate the effectiveness of our MMET method, we visualize some results of Grad-CAM and guided Grad-CAM~\citep{selvaraju2017grad} on Market-1501 dataset. As illustrated in Figure~\ref{fig4}, even though the position of the pedestrian in the whole image is relatively biased, our model can still accurately locate the pedestrians, which shows the better representation ability of our MMET model. Importantly, there exists obvious attention map in the backpack region of the pedestrian image. Consequently, our proposed method can pay more attention to the discriminative regions for feature representation learning, indicating that our MMET can greatly help the model to learn global context information and meaningful visual features with better semantic understanding.

\subsection{Discussion}
The MMET model is the first attempt to deal with generalizable Re-ID tasks from the perspective of multimodality. Based on the Multi-Modal Equivalent Transformer network in which image encoder and text encoder cooperate, it can effectively realize the decoupling of single-modal input and multi-modal input, which can significantly address the problem of incomplete modality. Despite its promising performance on person Re-ID, we note that there are several limitations in our MMET method.

\begin{table}[!t]
  \centering
  \caption{Performance comparison with between different pretraining datasets (RandPerson vs. FineGPR).}
  \setlength{\tabcolsep}{1.3mm}{
    \begin{tabular}{lcccccc}
    \toprule
    \multirow{2}[4]{*}{Methods}  & \multicolumn{2}{c}{Market-1501} & \multicolumn{2}{c}{DukeMTMC-reID} & \multicolumn{2}{c}{CUHK03} \\
\cmidrule{2-7}                & mAP   & Rank-1 & mAP   & Rank-1 & mAP   & Rank-1 \\
    \midrule
    RandPerson~\citep{wang2020surpassing}     & 28.8  & 55.6   & 27.1     & 47.6    & 10.8     & 13.4  \\
    FineGPR~\citep{xiang2021less}  &  28.8  &  59.2  &  23.5  &  43.6  &  20.3  &  23.6 \\
    \bottomrule
    \end{tabular}}%
  \label{tab4}%
\end{table}%

First, as illustrated in Table~\ref{tab4}, when tested on the DukeMTMC-reID dataset, we found an interesting phenomenon that performance of MMET on FineGPR~\citep{xiang2021less} is slightly inferior and less competitive when compared on backbone on RandPerson~\citep{wang2020surpassing} (Rank-1 \textcolor[rgb]{1.00,0.39,0.09}{\textbf{47.6\%}} vs. \textcolor[rgb]{0.20,0.40,0.80}{\textbf{43.6\%}}). Generally speaking, the effectiveness of Vision Transformer can be more obvious when training data of identity reaches a certain scale (\textit{e.g.} RandPerson contains 1,801,816 images of 8,000 identities). At present, the scale and diversity of FineGPR datset still have a certain distance compared with the data in the field of NLP task, which may hinder the further performance improvement of our method on generalizable Re-ID task.

Second, we note that images and languages are signals of a different nature and this difference must be addressed carefully.
In this work, we choose FineGPR as the training set since we wants to make full use of the fine-grained attributes through our proposed MMET framework. However, like all natural data, these data have biases, potentially affecting the performance of our models.  These issues warrant further research and consideration when building upon this work to learn robust visual-semantic embedding.

\section{Conclusion and Future Work}
\label{sec5}
In this work, we propose a multi-modal equivalent Transformer for robust visual-semantic embedding learning, and a dynamic masking mechanism is introduced to learn better representation for person Re-ID event, as well as a novel set of objectives to achieve this goal. Comprehensive experiments conducted on a wide variety of person Re-ID datasets prove the effectiveness of our method, proving itself as a strong baseline for visual-semantic embedding learning. Importantly, our work points the way forward towards generalized but open models that perform well on a wide variety of multimodal data. Future work includes explaining the decisions made by deep networks in domains such as reinforcement learning, natural language processing and video applications.

\bmhead{Acknowledgments}

This work was supported by the National Natural Science Foundation of China under Grant No. 61977045.
The authors would like to thank the anonymous reviewers for their valuable suggestions and constructive criticisms.

\section*{Declarations}

\begin{itemize}
\item \textbf{Funding} \\  This work was partially supported by the National Natural Science Foundation of China under Grant No. 61977045.
\item \textbf{Conflict of interest} \\  The authors declare that they have no conflict of interest.
\item \textbf{Ethics approval} \\  All procedures performed in studies involving human participants were in accordance with
the ethical standards of the institutional and/or national research committee.
\item \textbf{Consent to participate} \\  All human participants consented for participating in this study.
\item \textbf{Consent for publication} \\  All contents in this paper are consented for publication.
\item \textbf{Availability of data and material} \\  The data used for the experiments in this paper are available online, see Section~\ref{sec4.1} for more details.
\item \textbf{Code availability} \\  The Re-ID baseline implementation is open-source; The dataset is also publicly available at \url{https://github.com/JeremyXSC/MMET}.
\item \textbf{Authors' contributions} \\  Suncheng Xiang and Mengyuan Guan contributed conception and design of the study. Mengyuan Guan contributed to experimental process and evaluated and interpreted model results. Yuzhuo Fu obtained funding for the project. Jingsheng Gao, Jiacheng Ruan, Chengfeng Zhou, Ting Liu, Dahong Qian and Yuzhuo Fu provided clinical guidance. Suncheng Xiang drafted the manuscript. All authors contributed to manuscript revision, read and approved the submitted version.
\end{itemize}

%
%
%
%
%
%
%

\bibliography{sn-bibliography}


\end{document}